\documentclass{article}

    \PassOptionsToPackage{numbers, compress}{natbib}

\newcommand{\codelink}{https://github.com/gkwt/rbo}

    \usepackage[preprint]{neurips_2024}

\usepackage[T1]{fontenc}    
\usepackage{hyperref}       
\usepackage{url}            
\usepackage{booktabs}       
\usepackage{amsfonts}       
\usepackage{nicefrac}       
\usepackage{microtype}      
\usepackage{xcolor}         
\usepackage{multirow}
\usepackage{bm}
\usepackage{graphicx} 
\usepackage{soul}
\usepackage{color}
\usepackage{caption}
\usepackage{float}

\newcommand{\mycaption}[2]{\caption{\textbf{#1}. #2}}
\newcommand{\bftab}{\fontseries{b}\selectfont}

\title{Ranking over Regression for Bayesian Optimization and Molecule Selection}

\author{%
  Gary Tom\\
  University of Toronto\\
  Vector Institute for Artificial Intelligence\\
  Toronto, ON, Canada \\
  \texttt{gtom@cs.utoronto.ca}
  \And
  Stanley Lo \\
  University of Toronto\\
  Vector Institute for Artificial Intelligence\\
  Toronto, ON, Canada
  \And
  Samantha Corapi \\
  University of Toronto\\
  Vector Institute for Artificial Intelligence\\
  Toronto, ON, Canada
  \And
  Al\'an Aspuru-Guzik\thanks{CIFAR Lebovic Fellow, Canada 150 Chair, Director of Acceleration Consortium}\\
  University of Toronto\\
  Vector Institute for Artificial Intelligence\\
  Toronto, ON, Canada \\
  \texttt{alan@aspuru.com}
  \And
  Benjamin Sanchez-Lengeling \\
  University of Toronto\\
  Vector Institute for Artificial Intelligence\\
  Toronto, ON, Canada \\
  \texttt{ben.sanchez@utoronto.ca}
}

\begin{document}

\maketitle

\begin{abstract}
  Bayesian optimization (BO) has become an indispensable tool for autonomous decision-making across diverse applications from autonomous vehicle control to accelerated drug and materials discovery. With the growing interest in self-driving laboratories, BO of chemical systems is crucial for machine learning (ML) guided experimental planning. Typically, BO employs a regression surrogate model to predict the distribution of unseen parts of the search space. However, for the selection of molecules, picking the top candidates with respect to a distribution, the relative ordering of their properties may be more important than their exact values. In this paper, we introduce Rank-based Bayesian Optimization (RBO), which utilizes a ranking model as the surrogate. We present a comprehensive investigation of RBO's optimization performance compared to conventional BO on various chemical datasets. Our results demonstrate similar or improved optimization performance using ranking models, particularly for datasets with rough structure-property landscapes and activity cliffs. Furthermore, we observe a high correlation between the surrogate ranking ability and BO performance, and this ability is maintained even at early iterations of BO optimization when using ranking surrogate models. We conclude that RBO is an effective alternative to regression-based BO, especially for optimizing novel chemical compounds.

\end{abstract}

\section{Introduction}

In the age of increasing data complexity and expensive computations, Bayesian optimization (BO) \cite{mockus1974bayesian} emerges as a powerful tool to find optimal outcomes with minimal evaluations. By efficiently navigating complex search spaces, BO accelerates discovery and optimization across diverse scientific and engineering applications. In recent years, BO has been used in many applications in chemistry, such as in drug discovery \cite{graff2021accelerating, thompson2022optimizing, crivelli2023machine} and materials design \cite{gomez2018automatic, griffiths2020constrained, agarwal2021discovery, pollice2021data, griffiths2022data, zhang2019bayesian}, accelerating the rate of discovery for useful molecules, materials, and pharmaceuticals. Leveraging machine learning models, information extracted from available observations is extrapolated to unseen portions of the search space. This is particularly useful when the search space is extremely large \cite{graff2021accelerating, klarich2024thompson}, or when observations are expensive to evaluate and resource intensive, such as in high-quality chemical simulations, clinical trials, or laboratory experimentation. When coupled with automated high-throughput synthesis and characterization platforms, BO algorithms become a major component of experiment planning and decision-making in self-driving laboratories \cite{abolhasani2023rise, tom2024self}.

Underlying the machinery of BO, machine learning (ML) and deep learning (DL) regression models act as surrogates for complex quantitative structure-property relationships (QSPR) \cite{nigam2021assigning, walters2020applications, wu2018moleculenet}. Probabilistic models are typically preferred for the provided uncertainties for each prediction, and overall robustness to overfitting \cite{hirschfeld2020uncertainty, hwang2020benchmark, scalia2020evaluating}. An acquisition function dictates the strategy of exploration or exploitation based on the surrogate results, and the candidates are evaluated in the subsequent iteration. However, in the early stages of BO and scientific discovery there is often a limited amount of high-quality data which does not work well with DL regression models that struggle to learn the QSPR in the low-data regime, especially when compared to traditional ML methods, such as boosted random forest, and Gaussian processes \cite{tom2023calibration}. 

Additionally, when considering the optimization of the functional space of chemicals, the roughness of the space may present challenges to regression surrogate models. Extensive studies have been done looking at the roughness of chemical datasets, and the effect on the modellability of the QSPR using regression models \cite{peltason2007sar, guha2008structure, golbraikh2014data, aldeghi2022roughness, graff2023evaluating}. Increasing functional space roughness, and the presence of activity cliffs---small changes in the molecular structure corresponding to high fluctuations in the property---reduces the predictive accuracy of regression models \cite{maggiora2006outliers, van2022exposing}. Although smooth functional landscapes are best for regression modellability, many optimal compounds are often found at the activity cliffs \cite{stumpfe2020advances, dablander2023exploring, Lee2023-lm}, which are prevalent in many experimental data due to the complex physical effects that determine properties such as reactivity of a compound \cite{newman2021linking}, or the potency of a drug compound \cite{cruz2014activity}.

\begin{figure}[h]
    \centering
    \includegraphics[width=
    \textwidth]{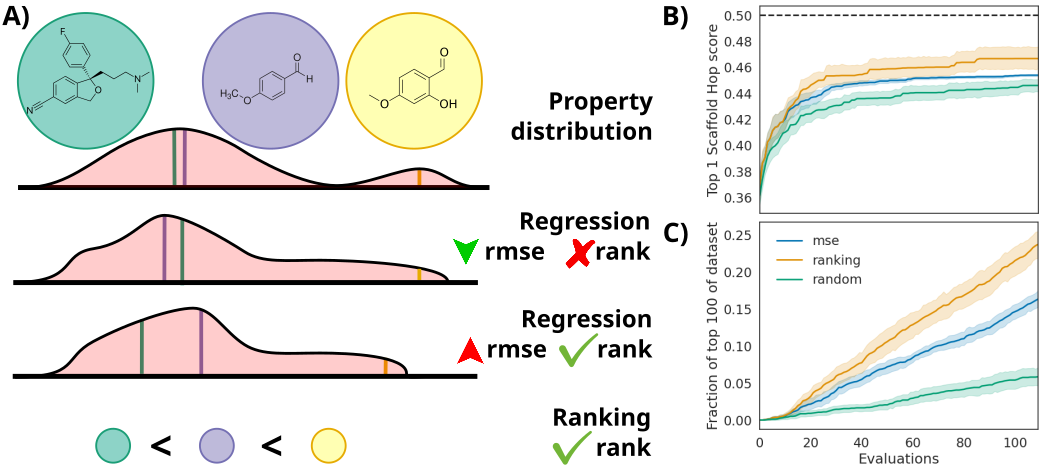}
    \mycaption{Schematic of regression and ranking models applied to modelling chemical systems}{A) Regression models aim to directly model the structure to functional space. Low errors in regression do not always lead to correct ranking of molecules, and can be sensitive to outliers. Pairwise ranking models directly learn the relative ranking of molecules. B) Ranking DL surrogate model can find higher scoring molecules, and C) find a greater proportion of population of top scorers.}
    \label{fig:problem}
\end{figure}

In this study, we consider the use of DL ranking models as an alternative to regression models for surrogates in BO. Unlike regression models, ranking models are trained to learn the relative ordering of candidates, rather than the exact target values \cite{liu2009learning, cao2007learning}. Ranking models have often been used in the context of efficient information retrieval, recommendation systems, and preference optimization, in which the most relevant results are prioritized \cite{guo2020deep, ko2022survey, nguyen2021top, gonzalez2017preferential}. In Rank-based Bayesian Optimization (RBO) for chemical systems, the surrogate model does not need to accurately learn the QSPR and instead becomes a molecule-selector. At the same time, the BO loop retrieves the most relevant molecules for further experimentation (Figure ~\ref{fig:problem}). In addition, the relative order of fitness of compounds are not affected by outliers and activity cliffs, effectively reducing sharp changes in the functional space for ranking models. 

To test the performance of the proposed RBO, we will simulate BO-guided molecular discovery on chemical datasets with varying roughnesses and compare the performance of the optimization between DL models trained with regression loss and ranking loss, as well as with traditionally used regression GP models. We will also compare the predictive and ranking abilities of the models throughout the campaign to the BO optimization performance. Our contributions and findings are summarized below:
\begin{itemize}\itemsep -2pt 
    \item Novel application of probabilistic ranking models in the BO of molecular datasets. All code and data provided here: \url{\codelink}
    \item Demonstrate similar or improved BO performance of rough chemical datasets with deep ranking models when compared with regression models. In some cases, deep ranking models outperform GP surrogates.
    \item Ranking performance is a more important metric of surrogate BO ability than regression performance.
    \item Ranking loss enables DL models to achieve better ranking performance than regression loss, particularly at low-data regimes early in the optimization.
\end{itemize}

\section{Methods}\label{sec:methods}

In RBO, we test the optimization performance of some common DL models as surrogates: multi-layer perceptron (MLP) \citep{bishop1994neural, anderson1995introduction}, bayesian neural network (BNN) \cite{blundell2015weight}, and graph neural network (GNN) \cite{battaglia2018relational}. Additionally, we compare the results with a Gaussian process (GP) \citep{williams2006gaussian} surrogate, commonly used for low-data BO. The molecules are represented using the Morgan extended-connectivity fingerprint (ECFP) representation \cite{rogers2010extended}, a 2048 dimensional bit-vector hashed from local structures of radius 3 in the molecular graph, as implemented in cheminformatics software \texttt{RDKit} \cite{Landrum2016RDKit2016_09_4}. For the GNN, the molecules are represented as graphs with atoms as nodes and bonds as edges, along with node and edge features as defined in the Open Graph Benchmark \citep{hu2020open}.

The GNN, BNN and the GP are all probabilistic models, while the MLP is deterministic. The GNN is based on the ChemProp architecture \citep{yang2019analyzing, heid2023chemprop}, with two message-passing layers operating and a final variational inference Bayesian layer to produce an uncertainty in the prediction \citep{krishnan2022bayesiantorch}. The same variational Bayesian layer is used to build the BNN, with two hidden layers each with 100 nodes and the rectified linear activation function. The MLP has the same architecture as the BNN, but is composed on fully connected feed-forward layers. The DL models are implemented in \texttt{PyTorch} \citep{paszke2019pytorch}, and the GNN in \texttt{PyTorch Geometric} \cite{fey2019fast, pygchemprop}. 

A GP defines a distribution over functions, characterized by a mean and covariance kernel function. Here, we use the Tanimoto distance kernel, which is effective when using Morgan fingerprint representations of molecules \cite{moss2020gaussian, tom2023calibration}. Uncertainty in predictions is inherent in GPs and is represented by the variance of the posterior distribution over predicted outputs. Unlike the deep Bayesian layers, the GP posterior is directly inferred, without variational inference or approximations. The GP is implemented using \texttt{GPyTorch} \cite{gardner2018gpytorch} and \texttt{GAUCHE} \cite{griffiths2024gauche}.

\subsection{Loss}\label{sec:loss}

For the regression models, the DL models are trained with a mean squared error (MSE) loss. The probabilistic GNN and BNN models have an additional regularization term from the KL divergence over the weight distributions of the variational layers. GPs are trained by minimizing the negative marginal log-likelihood (MLL).

The ranking loss used for the LTR task is the pairwise marginal ranking loss. Unlike a point-wise loss like MSE, which maps a scalar prediction and ground truth to a scalar loss value ($\mathbb{R} \times \mathbb{R} \rightarrow \mathbb{R}$), a pairwise loss maps a pair of predictions and a pair of ground truths to a scalar loss ($\mathbb{R}^2 \times \mathbb{R}^2 \rightarrow \mathbb{R}$). The pairwise marginal ranking loss has the form
\begin{equation}
    \mathcal{L}(y_1, y_2, \hat{y}_1, \hat{y}_2) = \max\big( 0, - \textrm{sign} (y_1 - y_2) * (\hat{y}_1 - \hat{y}_2) + m \big),
\end{equation}
where $(y_1, y_2)$ is the ground truth pair, $(\hat{y}_1, \hat{y}_2)$ is the predicted pair and $m$ is a margin for which the predicted ranks are allowed to overlap (here we allow for no margin, $m=0$). For correctly ranked predicted pairs, the sign of the second argument will be negative and $\mathcal{L} = 0$. Note that the ordering of the pairs does not matter, and will produce the same loss. During training, the dataset is collated into $(N^2-N)/2$ unique pair combinations, where $N$ is the dataset size. This can quickly become intractable, so $2N$ pairs are randomly sampled for the training set.

\subsection{Datasets}\label{sec:datasets}

The regression and ranking models are tested on the optimization of a variety of datasets. To test the robustness of optimization, we follow the steps of \citet{aldeghi2022roughness} in generating a series of datasets with varying degrees of roughness. Samples of 2000 molecules are taken from the ZINC 250k dataset \cite{gomez2018automatic}, a subset of the ZINC15 database \cite{sterling2015zinc} of drug-like molecules. The properties for optimization are derived from 12 oracles from the Therapeutics Data Commons (\texttt{PyTDC}) \cite{PyTDC, Huang2021tdc}, including 10 from the GuacaMol benchmark \cite{brown2019guacamol}, the LogP solubility measure \cite{wildman1999prediction}, and the Quantitative Estimate of Drug-likeness (QED) measure \cite{bickerton2012quantifying}. The roughness of the datasets is quantified by the Roughness Index (ROGI). ROGI measures how quickly the dispersion of a dataset changes as it is clustered by the pairwise distance between all molecules in the dataset across different distance thresholds \cite{bajusz2015tanimoto, lipkus1999proof}. Although there is no definitive metric for roughness, this index has been shown to correlate well with out-of-sample ML model error better than other existing indices (i.e. Structure-Activity Landscape Index \cite{guha2008structure}, Structure-Activity Relationship Index \cite{peltason2007sar}, Modellability Index \cite{golbraikh2014data}). Hence, we evaluate the RBO approach across chemistry datasets with varying roughness levels as measured by ROGI.

Additionally, we study the effects of activity cliffs on the BO performance. MoleculeACE \cite{van2022exposing} provides a curated set of datasets derived from the ChEMBL dataset (v29) \cite{gaulton2012chembl} with highly rugged functional space and activity cliffs. This is particularly important when considering BO of unexplored chemical space, or for optimization of novel characterization methods with highly complex relationships between feature space and functional space. The curated datasets involve the minimization of the inhibitor constant $K_i$ (the concentration required for half-maximal inhibition of a target protein) and the half-maximal effective concentration $\textrm{EC}_{50}$ (the concentration required for half-maximal response of the drug), which both measure the effectiveness of drug molecules.

For targets that are not already normalized, we perform additional scaling by subtracting the median value and dividing by the inter-quartile range of the distribution of targets. This is similar to standard scaling, but does not assume a normal distribution, and is more robust to outliers.

\begin{figure}[h]
    \centering
    \includegraphics[width=\textwidth]{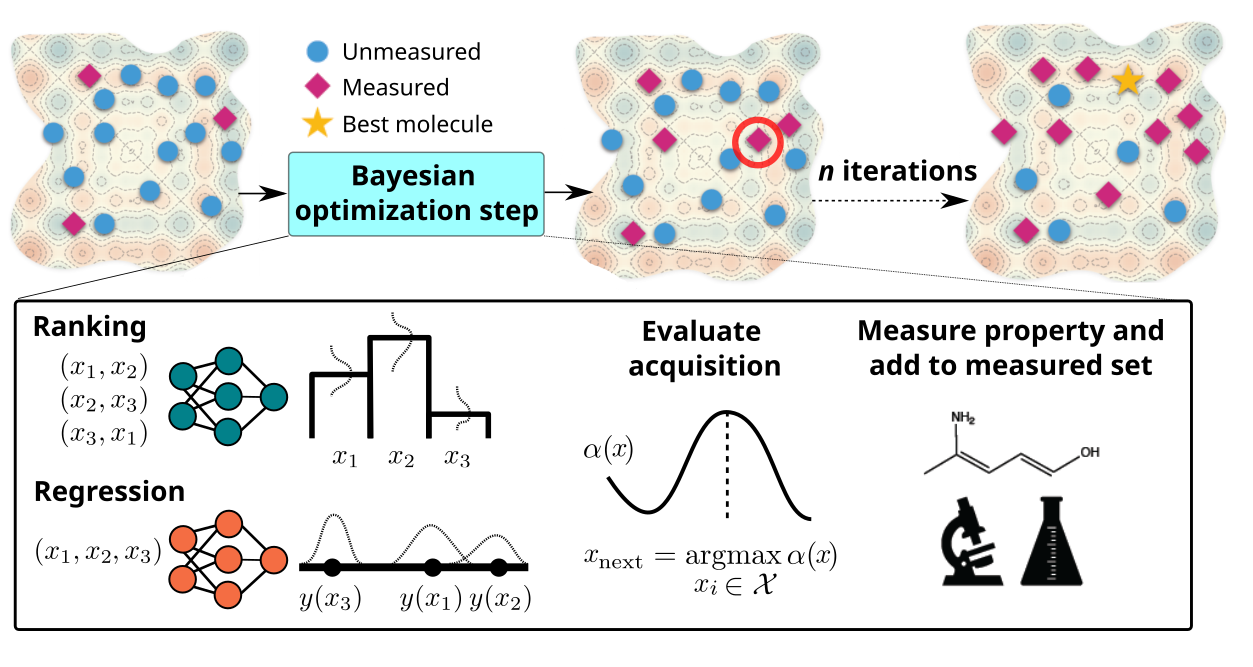}
    \mycaption{Schematic of experimental setup for BO campaigns}{In the BO step, either a pairwise ranking model or regression model is trained. Model inference produces the distribution of predicted ranking or property value on the unmeasured set, and the acquisition function is maximized. The selected molecules are then added to the measured set, simulating a property measurement on an unobserved molecule.}
    \label{fig:setup}
\end{figure}

\subsection{Bayesian optimization}\label{sec:bayesopt}
The different models are used as surrogates in Bayesian optimization of various chemical datasets (Figure \ref{fig:setup}) to find the molecule with the optimal property within the dataset with as few iterations as possible. The surrogates are trained on a subset (the measured set) of the dataset first, producing a series of predictions on the unseen part of the dataset. The candidates are then scored by an acquisition function and selected by $x_{next} = \arg\max \alpha(\hat{y}(x), \hat{\sigma}(x))$, where $\hat{y}$ and $\hat{\sigma}$ are the mean and standard deviation of the prediction on molecule $x$. 

For probabilistic models, the predictions include an uncertainty, which are incorporated into an acquisition function. Here, we study the optimization with the upper-confidence bound
\begin{equation}
    \alpha_{UCB}(\hat{y}, \hat{\sigma}) =  \hat{y} + \beta \hat{\sigma},
\end{equation}
where $\beta$ is a hyperparameter for controlling the exploration of the BO algorithm. We set $\beta = 0.3$. The MLP is deterministic and does not produce a posterior distribution. The predictions are maximized directly in a greedy approach. Results on the expected improvement acquisition function are shown in the Supplementary Materials.

The BO campaigns start with 10 initial randomly sampled data points, followed by 100 additional evaluations. The surrogate models are re-trained every 5 evaluations for up to 100 epochs with early stopping, with the validation loss monitored on a 90/10 training/validation random split of the measured set. The optimization uses the Adam algorithm \cite{KingBa15} with learning rate $0.001$ for the DL models, and $0.01$ for the GP. Statistics are obtained from 20 differently seeded BO runs. It takes at most 1 hour to complete 20 campaigns for a particular surrogate model on 1 GPU (NVIDIA GeForce RTX 2080), 4 CPUs (Intel Xeon Gold 5217), and 16GB of memory.

\section{Results}\label{sec:results}

To quantify the BO performance, we use the area under curve (AUC) of the fraction of top 100 molecules found as a function of BO evaluations, normalized by the budget. The BO-AUC metric rewards surrogate models that identify the most top-performing molecules in the earlier iterations of the campaign. Additionally, throughout the optimization campaigns, the surrogate model performance is evaluated on a held-out test set (15\% of the entire dataset). For this, we measure the performance of the model through the ranking Kendall tau correlation coefficient, and the $R^2$ coefficient of determination, at each BO iteration.

\subsection*{Ranking gives faster and better optimization}

\begin{table}[h]
\mycaption{BO performance of BNN and GNN surrogates}{Comparison of BO-AUC metrics for BO with probabilistic DL models trained with MSE and ranking loss on the ZINC datasets. Higher is better; the 95\% confidence interval is provided. Bold values are statistically significant. \\}
\label{tab:zinc_prob_auc}
\centering
\resizebox{\columnwidth}{!}{
\begin{tabular}{lcccccc}
\toprule
 & \multicolumn{2}{c}{BNN + UCB} & \multicolumn{2}{c}{GNN + UCB} & \multicolumn{2}{c}{MLP + greedy}   \\
\cmidrule(lr){2-3}\cmidrule(lr){4-5} \cmidrule(lr){6-7}
 & MSE & Ranking  & MSE  & Ranking & MSE  & Ranking  \\
\midrule
Amlodipine MPO \cite{brown2019guacamol} & 0.029 ± 0.080 & \underline{0.042 ± 0.008} & 0.032 ± 0.005 & \underline{0.037 ± 0.007} & 0.024 ± 0.005 & \bftab{0.044 ± 0.009} \\
Aripiprazole Similarity \cite{brown2019guacamol} & 0.088 ± 0.011 & \underline{0.098 ± 0.014} & 0.039 ± 0.006 & \bftab{0.067 ± 0.011} & 0.094 ± 0.007 & \underline{0.103 ± 0.012}  \\
Celecoxib Rediscovery \cite{brown2019guacamol} & 0.099 ± 0.010 & \bftab{0.136 ± 0.013} & 0.098 ± 0.008 & \bftab{0.208 ± 0.014} & 0.108 ± 0.009 & \bftab{0.149 ± 0.014} \\
Fexofenadine MPO \cite{brown2019guacamol} & \bftab{0.093 ± 0.010} & 0.067 ± 0.008 & 0.007 ± 0.005 & \bftab{0.047 ± 0.012} & \bftab{0.112 ± 0.009} & 0.065 ± 0.011 \\
LogP \cite{wildman1999prediction} & \underline{0.116 ± 0.014} & 0.106 ± 0.012 & \underline{0.240 ± 0.010} & 0.232 ± 0.013 & \underline{0.118 ± 0.011} & 0.100 ± 0.012 \\
Median 1 \cite{brown2019guacamol} & 0.051 ± 0.017 & \bftab{0.135 ± 0.030} & 0.269 ± 0.018 & \bftab{0.436 ± 0.009} & 0.068 ± 0.019 & \bftab{0.130 ± 0.025} \\
Osimertinib MPO \cite{brown2019guacamol} & \underline{0.049 ± 0.007} & 0.046 ± 0.008 & 0.018 ± 0.004 & \bftab{0.110 ± 0.009} & \underline{0.053 ± 0.005} & 0.052 ± 0.007 \\
Perindopril MPO \cite{brown2019guacamol} & \underline{0.046 ± 0.009} & 0.038 ± 0.009 & 0.008 ± 0.003 & \bftab{0.050 ± 0.015} & \underline{0.043 ± 0.009} & 0.041 ± 0.010 \\
QED \cite{bickerton2012quantifying} & 0.020 ± 0.006 & \bftab{0.035 ± 0.006} & 0.043 ± 0.005 & \underline{0.053 ± 0.010} & 0.018 ± 0.005 & \bftab{0.035 ± 0.007}  \\
Ranolazine MPO \cite{brown2019guacamol} & 0.077 ± 0.012 & \underline{0.084 ± 0.012} & 0.080 ± 0.012 & \bftab{0.112 ± 0.012} & \underline{0.092 ± 0.007} & 0.086 ± 0.012 \\
Scaffold Hop \cite{brown2019guacamol} & 0.075 ± 0.008 & \bftab{0.113 ± 0.010} & 0.137 ± 0.010 & \bftab{0.297 ± 0.011} & 0.076 ± 0.007 & \bftab{0.113 ± 0.013}  \\
Zaleplon MPO \cite{brown2019guacamol} & 0.041 ± 0.006 & \bftab{0.056 ± 0.007} & 0.099 ± 0.012 & \bftab{0.139 ± 0.011} & 0.050 ± 0.007 & \underline{0.053 ± 0.008} \\
\bottomrule
\end{tabular}}
\end{table}

In Table \ref{tab:zinc_prob_auc}, the BO-AUC metrics on the optimization of DL models for the ZINC datasets are shown for regression loss (MSE), and ranking loss (pairwise margin ranking loss). For almost all targets, the surrogate trained with the ranking loss achieves higher BO-AUC, indicating faster and better BO. Solely on the Fexofenadine multi-property objective (MPO), the feed-forward network regressors outperform their ranking counterparts.

Notably, the GNN ranking models dramatically improve upon the GNN regression models. While the ChemProp message-passing architecture typically achieves state-of-the-art performance on regression tasks, the GNN regression model struggles in the low-data regime making them poor surrogates for BO of molecules. However, for some datasets, the GNN ranking models achieve optimization results competitive to or better than the feed-forward DL networks. For example, on the Median 1 task, the GNN ranking model achieves an average BO-AUC score of 0.4355, more than three times the BO-AUC of the BNN (0.1351) and MLP (0.1303) ranking models.


\subsection*{The effect of dataset roughness on ranking and regression}

\begin{figure}[h]
    \centering
    \includegraphics[width=\textwidth]{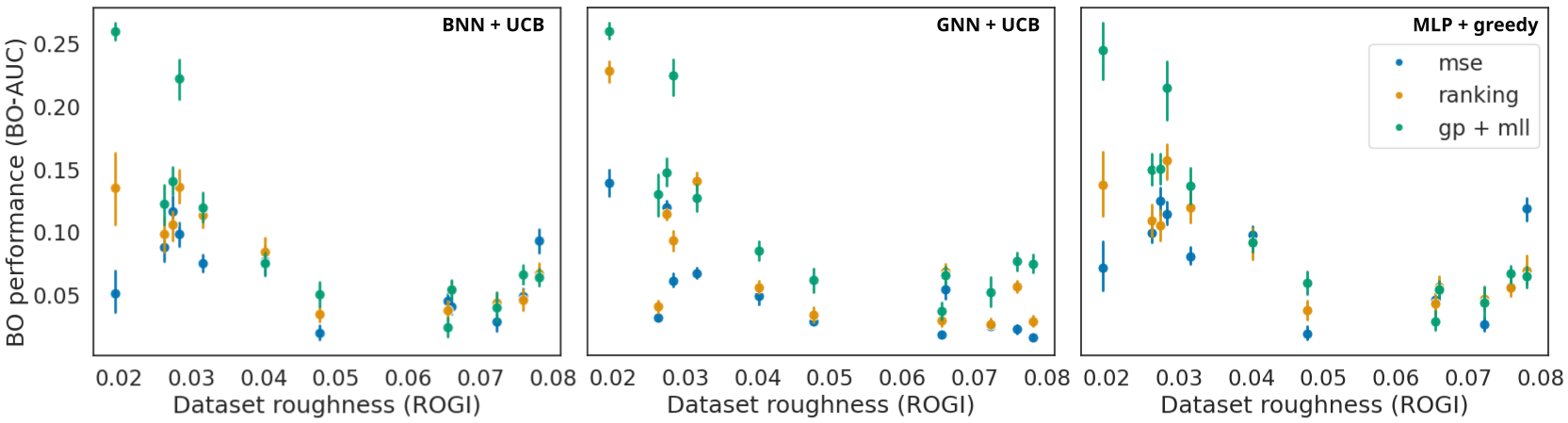}
    \mycaption{BO performance compared to ROGI of ZINC datasets}{The BO-AUC values correspond to those of Table \ref{tab:zinc_prob_auc}. Here the values are visualized as a function of the ROGI of the datasets. Rougher datasets have higher ROGI. The GP surrogate results are shown in all plots with the respective acquisition functions.}
    \label{fig:rogi_v_opt}
\end{figure}

Following the procedures of \citet{aldeghi2022roughness}, the BO performance can be correlated to the roughness of the dataset using the ROGI score. Higher ROGI scores correspond to rougher datasets. The constructed datasets are from the same chemical space, sampled from ZINC, and contain the same number of data points, making the comparison fair across the different properties. The BO-AUC of the BO campaigns compared to the ROGI of the ZINC datasets are shown in Figure ~\ref{fig:rogi_v_opt}, with additional comparison to the performance of a GP regression surrogate. We see a general trend of decreasing optimization performance with increasing ROGI for all surrogates. As observed before, the ranking models often perform better than the DL regression models. The optimization performance of the GP models suffers on datasets with higher ROGI, with DL models occasionally achieving similar BO-AUC. Still, for smoother datasets, the GPs are the surrogates of choice.

We also compare the relative performances of the BNN, GNN, MLP and GP surrogates on the MoleculeACE datasets by performing the Student's t-test on the achieved BO-AUC metrics (Figure ~\ref{fig:ki_ec50}). Individual BO-metrics achieved on each dataset is provided in the Supplementary Materials. On average, the DL ranking models serve as better surrogates when compared to the same model trained with MSE loss. While the GNN model struggles as a surrogate for all MoleculeACE datasets when compared to the GP model, the ranking model still readily improves upon the regression model. In the original benchmarking work of MoleculeACE, \citet{van2022exposing} studied the regression of the same datasets using DL models, and found that GNNs had the worst performance on activity cliffs, while MLPs had the best performance. This corresponds well with our findings on the poor BO performance of GNN surrogates on the MoleculeACE datasets.

For the feed-forward BNN and MLP models, BO performance is similar to, if not better than, GP surrogates, for all MoleculeACE datasets. This is surprising, as traditional ML methods such as random forest and GPs are typically preferred over DL methods, which tend to be over-parameterized and overfit in the low-data regimes. This suggests that GPs struggle with the prediction of activity cliffs. However, between ranking and regression BNN and MLP, ranking model is still the model of choice, offering better performance on more datasets than both GPs and their regression counterparts. Similar to the results observed on the ZINC datasets, ranking models are more robust to rough functional space, activity cliffs, and outliers. 

\begin{figure}[h]
    \centering
    \includegraphics[width=\textwidth]{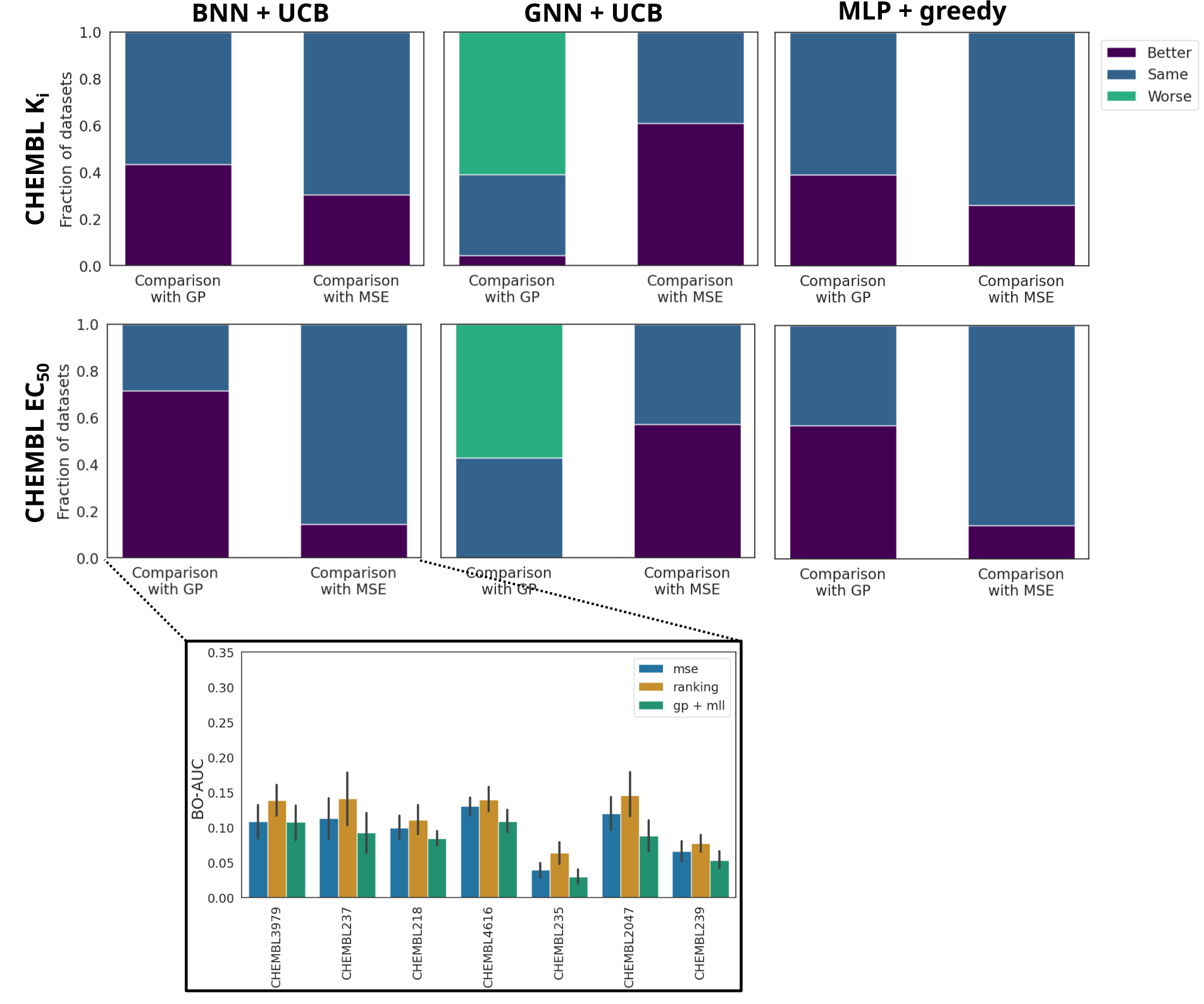}
    \mycaption{Comparison of BO-AUC achieved by ranking models versus GPs and regression models for BO of MoleculeACE datasets}{The bar plots show the fraction of datasets from ChEMBL $K_i$ and $\textrm{EC}_{50}$ that the ranking model achieved better, similar, or worse BO-AUC, based on the Student's t-test ($p < 0.05$). This information is summarized as a statistic over all the datasets within the given property (7 datasets with $\textrm{EC}_{50}$, and 23 datasets with $K_i$), and for one subplot, the full results are shown.}
    \label{fig:ki_ec50}
\end{figure}

\subsection*{Ranking performance is better correlated to better selection}

To elucidate the correlations between BO performance and the surrogate models' predictive and ranking abilities, the models are evaluated on a held-out test set at each BO iteration. The Pearson correlation coefficient is used to examine the relationship across the datasets between the fraction of top 100 molecules observed at the end of the campaigns and the final surrogate test performance. For regressive performance, even for the models trained with the MSE loss and the GP regressor, the $R^2$ metric has a tenuous relationship to the BO performance, with the Pearson test failing to produce any correlations with $p < 0.05$.

\begin{table}[h]
\mycaption{Correlation of test Kendall tau metric with BO performance}{Table of Pearson correlation metrics across the various datasets for observed surrogate test performance and the maximum fraction of top 100 molecules found. Only correlations with $p < 0.05$ are reported. \\}
\label{tab:test_corr}
\centering
\begin{tabular}{lccccccc}
\toprule
& \multicolumn{2}{c}{BNN  + UCB} & \multicolumn{2}{c}{GNN + UCB} & \multicolumn{2}{c}{MLP + Greedy}  & GP + UCB    \\
\cmidrule(lr){2-3}\cmidrule(lr){4-5}\cmidrule(lr){6-7}\cmidrule(lr){8-8}
Dataset & MSE & Ranking  & MSE  & Ranking  & MSE & Ranking & MLL \\
\midrule
ZINC & 0.826 & 0.917 & 0.918 & 0.796 & 0.846  & 0.919 &  0.905 \\
ChEMBL $K_i$ & 0.501 & 0.555 &- & - & 0.436 & 0.522 & 0.609 \\
ChEMBL $\textrm{EC}_{50}$ & 0.660 & 0.717 & 0.839 & 0.802 & 0.487 & 0.616 & 0.537 \\
\bottomrule
\end{tabular} 
\end{table}


Conversely, we find a positive correlation in the model ranking performance, evaluated by the test Kendall tau statistic, and the BO performance for both the regression and ranking surrogate models (Table \ref{tab:test_corr}). Stronger correlations are found across the ZINC datasets, while the MoleculeACE datasets exhibit lower, but positive correlation, likely an effect of the activity cliffs in these datasets. Additionally, with the exception of the GNN models on the ZINC datasets, the ranking surrogate models typically achieve a higher Pearson correlation between the test Kendall tau and the BO performance indicator. We conclude that the ranking performance of the surrogate model is a stronger indicator of performance as a BO surrogate.

\begin{figure}[!ht]
    \centering
    \includegraphics[width=0.8\textwidth]{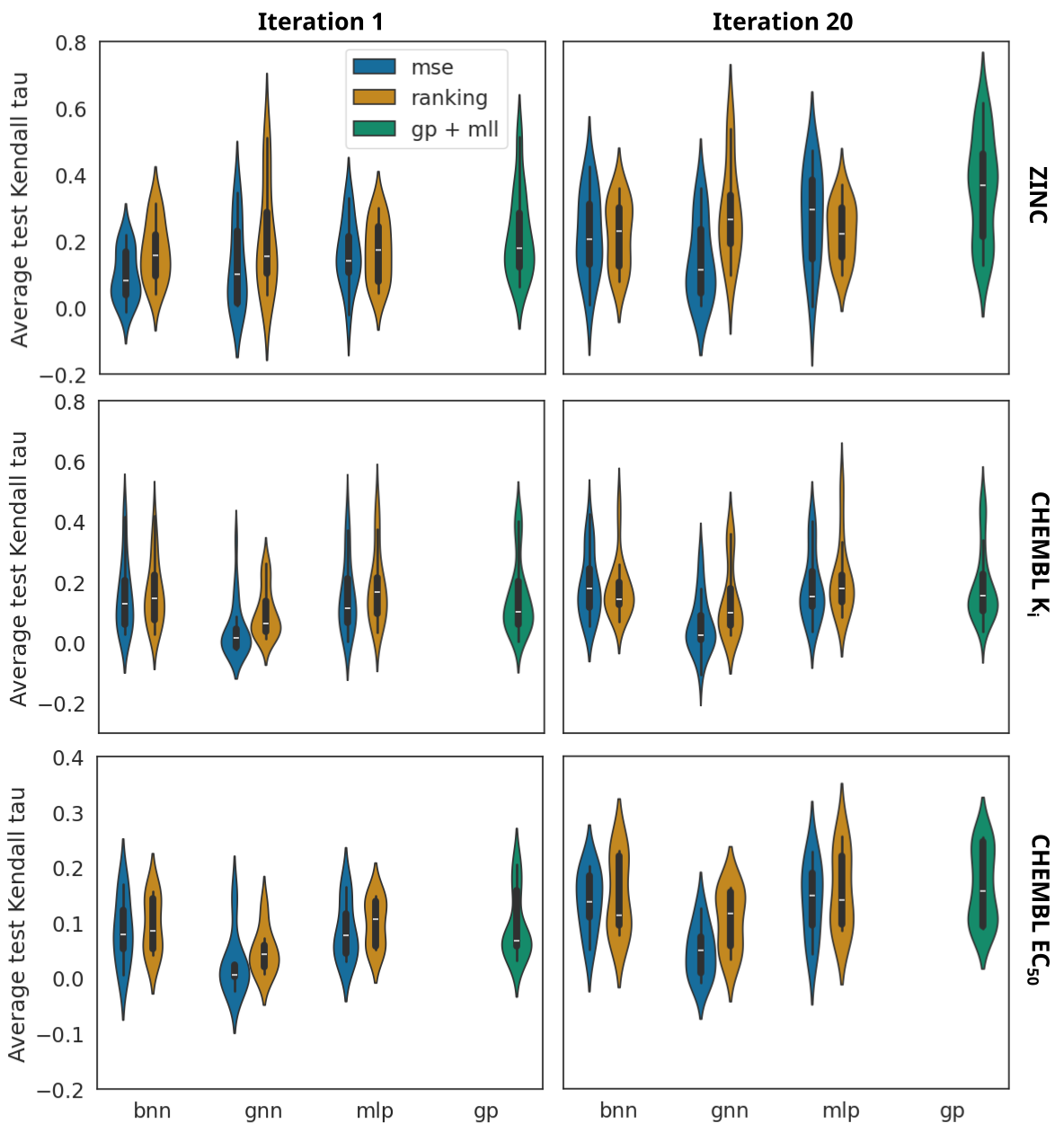}
    \mycaption{Violin plots of the test Kendall tau statistic over the ZINC and ChEMBL datasets.}{Comparison of BNN, GNN, MLP, and GP surrogate ranking performance for each set of datasets. The left column are from the first iteration of the BO optimization, while the right column are from the last iteration.}
    \label{fig:violin}
\end{figure}

We further examine the surrogate model test ranking performance at two different BO iterations, averaged across the three sets of datasets (Figure ~\ref{fig:violin}). As expected, the average test ranking statistic is typically higher for the deep ranking models than the regression models. Notably, even in the low-data regime of the first iteration, ranking models achieve higher and positive test Kendall tau statistics, suggesting that the ranking performance of deep models trained on pairwise ranking loss is more robust to sparse-data conditions. Paired with our earlier finding of a positive correlation between the test Kendall tau statistic and the BO performance, this indicates that ranking models perform better in earlier iterations, allowing the model to identify top scoring molecules faster. In the final iteration 20 of the BO campaign, we see that the regression models can achieve ranking performance more competitive with the ranking models. 

\section{Discussion}\label{sec:discussion}

\subsection*{Limitations and open questions}

Despite the promising results, there are several limitations and open questions that warrant further investigation:
\begin{itemize} \itemsep -2pt
    \item Scope: Our study looks at synthetically generated datasets from ZINC with a diverse and well-utilized set of models and representations, aiming for a fair comparison. Model performance can change with carefully selected hyperparameters and alternative representations. More works needs to be done on larger datasets and non-synthetic properties.
    \item Training complexity and other ranking losses: Ranking models require pairwise comparisons, which can increase training time. Future research could explore more efficient training methods, such as list-wise comparisons \cite{jagerman2020accelerated, jagerman2022rax}. There are also a variety of other ranking losses that can be studied.
    \item Multi-task models: RBO leverages the ordinal nature of continuous numbers, which may not be the best strategy when considering multi-task optimization problems. Further study can be done on the optimization of multiple objectives through other scalarizations \cite{kusanda2022assessing, hickman2023olympus}, or pareto-optimal methods \cite{lin2022pareto, daulton2020differentiable, daulton2021parallel}.
    \item Broader applicability: While RBO showed success in the specific context of molecular selection, its applicability to other domains and types of data should be explored. For example, some success has been demonstrated in BO of ML model hyperparameter searches \cite{khazi2023deep}.
    \item Uncertainty quantification: We do not study the calibration of uncertainties \cite{guo2017calibration, minderer2021revisiting}, which is important to BO performance \cite{snoek2012practical, tom2023calibration, foldager2023role}. There is less work on estimating uncertainty of rank scores \cite{dewancker2018sequential}. We also do not utilize experimental error of the properties which could be used to better evaluate performance of models.
    \item Comparison to other selection strategies: a more comprehensive study can look at ranking in the context of other optimization strategies.
\end{itemize}

\subsection*{Conclusion}
This study introduces RBO as an alternative to traditional regression-based BO methods, particularly in the context of molecular selection. Our investigation demonstrates that RBO can either match or outperform conventional BO surrogates, especially in environments characterized by noisy or rough data.

The comparative analysis between regression and ranking models revealed that ranking models offer significant advantages in prioritizing high-performing candidates. Specifically, the ranking models showed superior performance on datasets with rough structure-property landscapes and activity cliffs, where the exact predictive values are less critical than the relative ordering of candidates. This advantage is attributed to the ranking models' robustness against outliers and noise, which often impair the performance of regression models.

Our findings suggest that the correlation between surrogate model performance and BO success is stronger for ranking models. The Pearson correlation analysis indicated a higher correlation between the test set's Kendall tau statistic and the BO performance for ranking models compared to regression models. This trend was consistent across various datasets, including ZINC and MoleculeACE, highlighting the efficacy of ranking models in identifying top-performing molecules early in the optimization process. By focusing on the relative ordering of candidate properties, RBO addresses key limitations of traditional regression-based models. Our results suggest that RBO is particularly effective in optimization of molecular datasets, making it a valuable tool for applications such as drug discovery and materials science.

In conclusion, RBO offers an alternative to traditional BO approaches, particularly in scenarios where the relative ranking of candidates is more critical than their exact predictive values. Future research should aim to address the identified limitations and expand the applicability of RBO to a wider range of optimization problems.

\begin{ack}
G.T. acknowledges the support of the Natural Sciences and Engineering Research Council (NSERC) of Canada, and the Vector Institute for Artificial Intelligence. S.L. acknowledges the support from the Government of Ontario through the Ontario Graduate Scholarship. This project was also part of the Acceleration Consortium Bayesian optimization hackathon. A.A.-G. acknowledges the generous support of Anders G. Fr\o seth, the CIFAR, and the Canada 150 Research Chair program. A.A.-G. is a founder of Kebotix, Inc., a company specializing in closed-loop molecular discovery, and Intrepid Labs, Inc. a company using self-driving laboratories for pharmaceuticals.
\end{ack}

\bibliography{neurips_2024}

\appendix

\section{Supplementary Materials}

\subsection{Results on ZINC datasets with Expected Improvement}

Additional studies were performed with the expected improvement (EI) acquisition function. The EI is defined as
\begin{equation}
    \alpha_{EI}(\hat{y}, \hat{\sigma}) = (\hat{y} - y^*) \Phi\left( \frac{\hat{y} - y^*}{\hat{\sigma}} \right) + \hat{\sigma} * \phi\left( \frac{\hat{y} - y^*}{\hat{\sigma}} \right),
\end{equation}
where $y^*$ is the best property found in the training set, and $\Phi$ and $\phi$ are the normal cumulative distribution function and probability density function, respectively. The EI scores the predictions by the probability of the point improving upon the observed property, based on a Gaussian distribution around the prediction. The EI takes on high values when $\hat{y} > y^*$, but also when $\hat{\sigma}$ is high. 

A table for the EI optimizations, analogous to Table \ref{tab:zinc_prob_auc}, is provided in Table \ref{tab:suppl_ei_rogi}.

\begin{table}[h]
\mycaption{BO performance of BNN, GNN, and GP surrogates using EI acquisition function}{Comparison of BO-AUC metrics for BO with probabilistic DL models trained with MSE and ranking loss on the ZINC datasets. Higher is better; the 95\% confidence interval is provided. Bold values are statistically significant. \\}
\label{tab:suppl_ei_rogi}
\centering
\resizebox{\columnwidth}{!}{
\begin{tabular}{lccccc}
\toprule
 & \multicolumn{2}{c}{BNN + EI} & \multicolumn{2}{c}{GNN + EI} & {GP + EI}   \\
\cmidrule(lr){2-3}\cmidrule(lr){4-5} \cmidrule(lr){6-6}
 & MSE & Ranking  & MSE  & Ranking &  MLL  \\
\midrule
Amlodipine MPO & 0.0290 ± 0.0088 & \underline{0.0395 ± 0.0103} & \underline{0.0145 ± 0.0032} & 0.0133 ± 0.0045& 0.0335 ± 0.0110 \\
Aripiprazole Similarity & 0.0921 ± 0.0092 & \underline{0.1081 ± 0.0113} & 0.0208 ± 0.0032 & \underline{0.0244 ± 0.0050} & 0.0922 ± 0.0137\\
Celecoxib Rediscovery & 0.0948 ± 0.0139 & \bftab{0.1323 ± 0.0149} & 0.0518 ± 0.0055 & \bftab{0.0890 ± 0.0102} & 0.1583 ± 0.0155 \\
Fexofenadine MPO & \underline{0.0929 ± 0.0103} & 0.0815 ± 0.0082 &   0.0030 ± 0.0017 & \bftab{0.0140 ± 0.0052}  & 0.0627 ± 0.0048 \\
LogP & \underline{0.0796 ± 0.0106} & 0.0651 ± 0.0110 & 0.0817 ± 0.0090 & \underline{0.0921 ± 0.0060} & 0.1574 ± 0.0066   \\
Median 1 & 0.0485 ± 0.0100 & \bftab{0.1106 ± 0.0296} & 0.1100 ± 0.0105 & \bftab{0.2106 ± 0.0113} & 0.2006 ± 0.0042 \\
Osimertinib MPO & \underline{0.0494 ± 0.0067} & 0.0493 ± 0.0088 & 0.0088 ± 0.0028 & \bftab{0.0338 ± 0.0044} & 0.0604 ± 0.0071 \\
Perindopril MPO & \underline{0.0452 ± 0.0099} & 0.0428 ± 0.0095 &  0.0047 ± 0.0025 & \bftab{0.0153 ± 0.0054}  & 0.0207 ± 0.0039 \\
QED & 0.0196 ± 0.0052 & \bftab{0.0327 ± 0.0053} & 0.0179 ± 0.0033 & \underline{0.0193 ± 0.0033} & 0.0399 ± 0.0083  \\
Ranolazine MPO & 0.0699 ± 0.0081 & \bftab{0.0919 ± 0.0095} & \underline{0.0425 ± 0.0068} & 0.0397 ± 0.0058  & 0.0524 ± 0.0069\\
Scaffold Hop & 0.0752 ± 0.0080 & \bftab{0.1140 ± 0.0096} &  0.0566 ± 0.0052 & \bftab{0.1203 ± 0.0075}   & 0.0695 ± 0.0068\\
Zaleplon MPO & 0.0495 ± 0.0066 & \underline{0.0571 ± 0.0056} & 0.0453 ± 0.0073 & \underline{0.0531 ± 0.0071} & 0.0431 ± 0.0094   \\
\bottomrule
\end{tabular}}
\end{table}

\subsection{Results on MoleculeACE datasets with EI}

\begin{figure}[h]
    \centering
    \includegraphics[width=\textwidth]{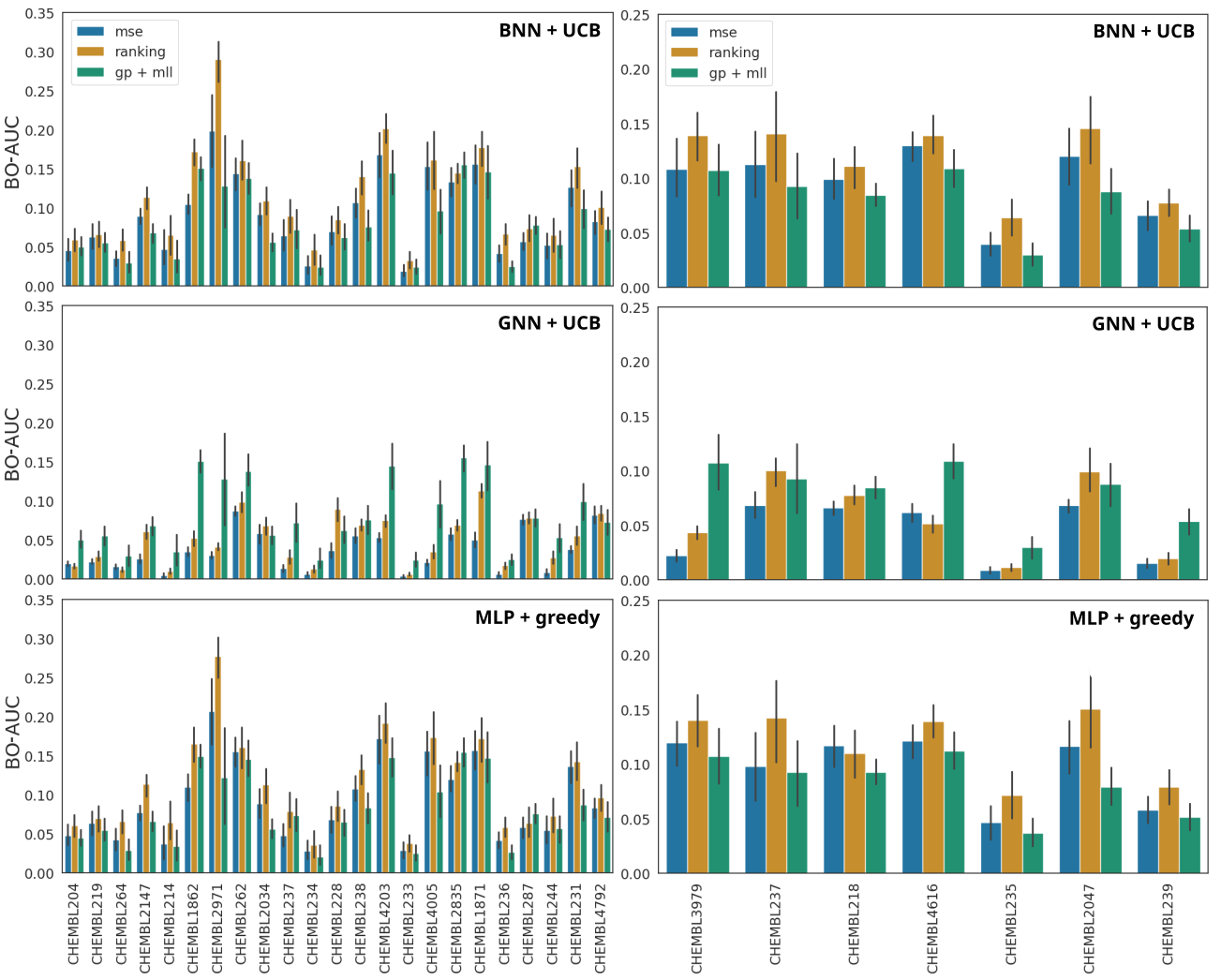}
    \mycaption{AUC metric of BNN, GNN, MLP, and GP surrogates for BO of MoleculeACE datasets}{Plots in the left column are the results on the $K_i$ datasets, and the right column the ChEMBL $\textrm{EC}_{50}$ datasets. The error bars represent the 95\% confidence interval. The GP surrogate results are shown in all plots with the respective acquisition functions.}
    \label{fig:ki_ec50_full}
\end{figure}

Full results have quite a lot of information, and so the summary of statistically significant results are displayed in Figure \ref{fig:ki_ec50}. Here the full results include the BO-AUC for all the datasets in MoleculeACE (Figure \ref{fig:ki_ec50_full}).

\begin{figure}[h]
    \centering
    \includegraphics[width=\textwidth]{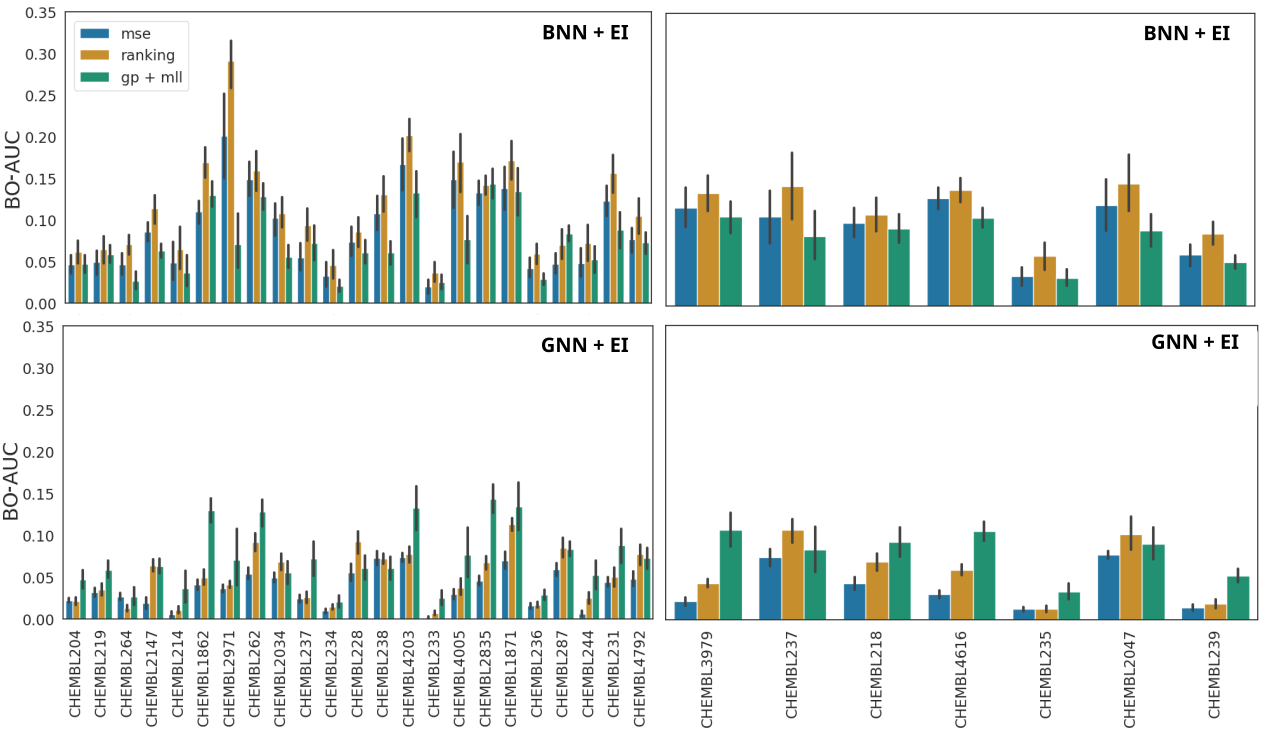}
    \mycaption{AUC metric of BNN, GNN, MLP, and GP surrogates for BO of MoleculeACE datasets}{Plots in the left column are the results on the $K_i$ datasets, and the right column the ChEMBL $\textrm{EC}_{50}$ datasets. The error bars represent the 95\% confidence interval. The GP surrogate results are shown in all plots with the respective acquisition functions.}
    \label{fig:ki_ec50_ei_full}
\end{figure}

Full results of BO with the EI acquisition function for all the datasets are shown in Figure \ref{fig:ki_ec50_ei_full}.

\clearpage

\begin{figure}[h]
    \centering
    \includegraphics[width=\textwidth]{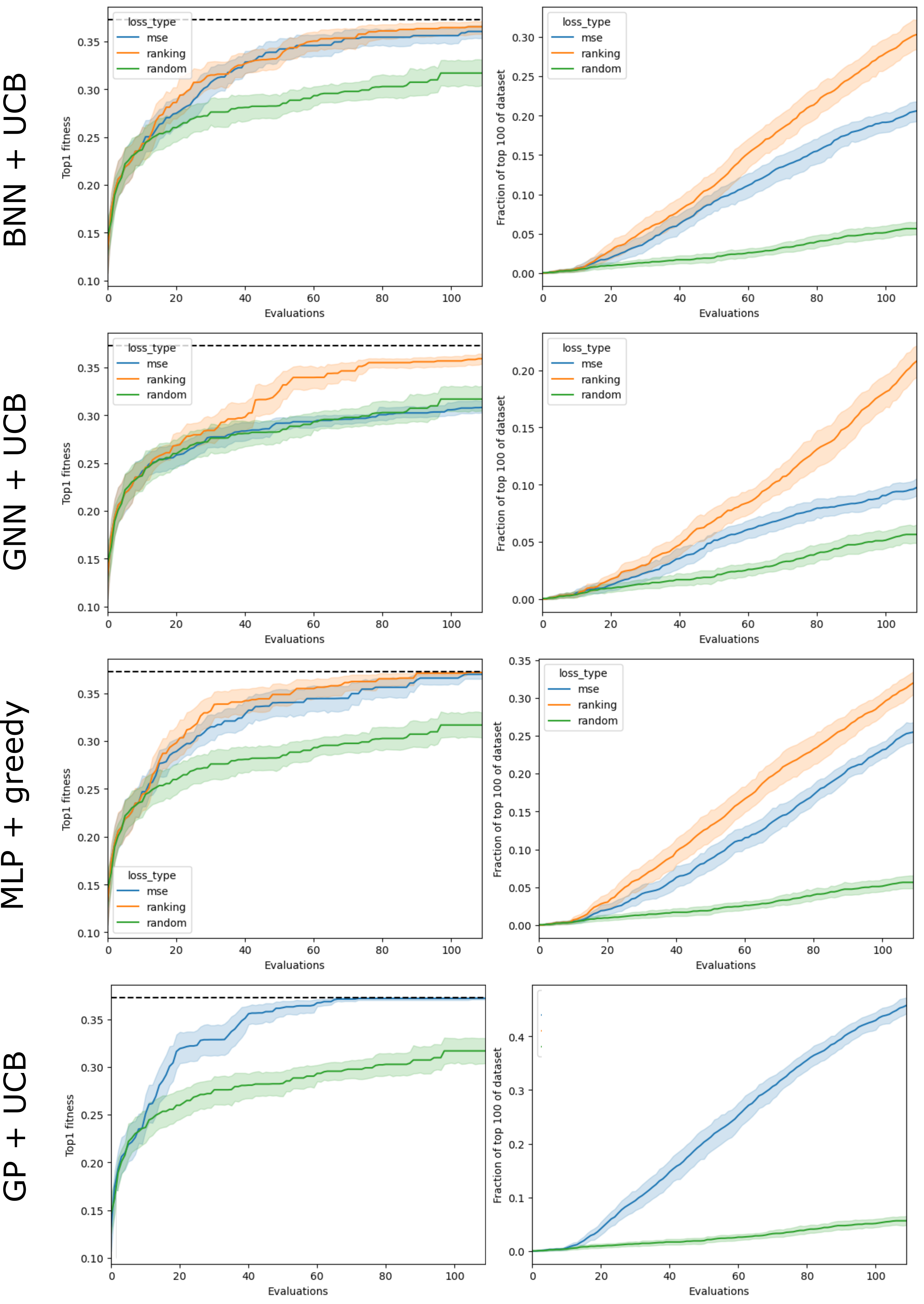}
    \mycaption{Optimization traces for Celecoxib rediscovery}{Traces of top-1 found and fraction of top-100 found.}
\end{figure}

\clearpage

\end{document}